\documentclass[10pt,twocolumn,letterpaper]{article}

\usepackage{iccv}
\usepackage{times}
\usepackage{epsfig}
\usepackage{graphicx}
\usepackage{amsmath}
\usepackage{amssymb}

\usepackage{multirow}
\usepackage{array}
\usepackage{booktabs}
\usepackage[table]{xcolor}
\usepackage[symbol]{footmisc}
\usepackage{url}
\usepackage[breaklinks=true,bookmarks=false]{hyperref}

\iccvfinalcopy 

\def\iccvPaperID{****} 

\ificcvfinal\fi

\begin{document}

\title{advPattern: Physical-World Attacks on Deep Person Re-Identification via Adversarially Transformable Patterns}

\author{Zhibo Wang$^{\dagger}$, Siyan Zheng$^{\dagger}$, Mengkai Song$^{\dagger}$, Qian Wang$^{\dagger,\ast}$,\footnotetext[1], Alireza Rahimpour$^{\ddagger}$, Hairong Qi$^{\ddagger}$\\
$^{\dagger}$Key Laboratory of Aerospace Information Security and Trusted
Computing, Ministry of Education,\\School of Cyber Science and Engineering, Wuhan University, P. R. China\\
$^{\ddagger}$Dept. of Electrical Engineering and Computer Science, University of Tennessee, Knoxville, USA\\
{\tt\small \{zbwang, zhengsy, mksong, qianwang\}@whu.edu.cn, \{arahimpo, hqi\}@utk.edu}
}

\maketitle
\footnotetext[1]{Qian Wang is the corresponding author.}
\ificcvfinal\thispagestyle{empty}\fi

\begin{abstract}
   Person re-identification (re-ID) is the task of matching person images across camera views, which plays an important role in surveillance and security applications. Inspired by great progress of deep learning, deep re-ID models began to be popular and gained state-of-the-art performance. However, recent works found that deep neural networks (DNNs) are vulnerable to adversarial examples, posing potential threats to DNNs based applications. This phenomenon throws a serious question about whether deep re-ID based systems are vulnerable to adversarial attacks.

  In this paper, we take the first attempt to implement robust physical-world attacks against deep re-ID. We propose a novel attack algorithm, called advPattern, for generating adversarial patterns on clothes, which learns the variations of image pairs across cameras to pull closer the image features from the same camera, while pushing features from different cameras farther. By wearing our crafted ``invisible cloak'', an adversary can evade person search, or impersonate a target person to fool deep re-ID models in physical world. We evaluate the effectiveness of our transformable patterns on adversaries' clothes with Market1501 and our established PRCS dataset. The experimental results show that the rank-1 accuracy of re-ID models for matching the adversary decreases from 87.9\% to 27.1\% under Evading Attack. Furthermore, the adversary can impersonate a target person with 47.1\% rank-1 accuracy and 67.9\% mAP under Impersonation Attack. The results demonstrate that deep re-ID systems are vulnerable to our physical attacks.
\end{abstract}

\section{Introduction}
\footnote{This work was accepted by IEEE ICCV 2019.}
Person re-identification (re-ID) \cite{gong2014re} is an image retrieval problem that aims at matching a person of interest across multiple non-overlapping camera views. It has been increasingly popular in research area and has broad applications in video surveillance and security, such as searching suspects and missing people \cite{wang2013intelligent}, cross-camera pedestrian tracking \cite{yu2013harry}, and activity analysis \cite{loy2009multi}. Recently, inspired by the success of deep learning in various vision tasks \cite{he2016deep,krizhevsky2012imagenet,simonyan2014very,szegedy2015going,zhang2017age,zhang2019image}, deep neural networks (DNNs) based re-ID models \cite{ahmed2015improved,chen2016deep,chen2017multi,cheng2016person,ding2015deep,li2014deepreid,wang2016joint,xiao2016learning,yi2014deep} started to become a prevailing trend and have achieved state-of-the-art performance. Existing deep re-ID methods usually solve re-ID as a classification task \cite{ahmed2015improved,li2014deepreid,yi2014deep}, or a ranking task \cite{chen2016deep,cheng2016person,ding2015deep}, or both \cite{chen2017multi,wang2016joint}.

\begin{figure}[!t]
  \centering
      \includegraphics[width=0.6\columnwidth]{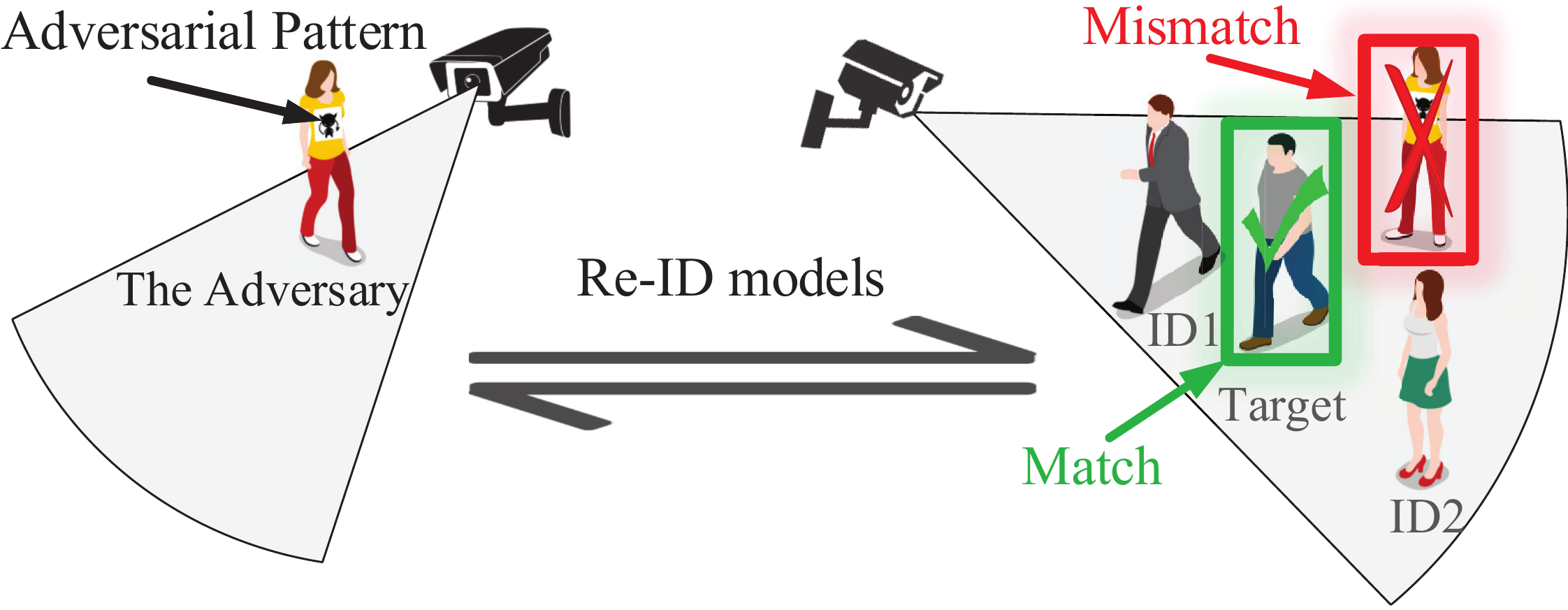}
  \caption{The illustration of Impersonation Attack on re-ID models. The adversary with the adversarial patterns lures re-ID models into mismatching herself as the target person.}
  \label{fig:example_advPattern}
  \vspace{-4mm}
\end{figure}

Recent studies found that DNNs are vulnerable to adversarial attack \cite{carlini2017towards,goodfellow6572explaining,kos2018adversarial,li2014feature,li2015scalable,moosavi2016deepfool,papernot2016limitations,szegedy2013intriguing}. These carefully modified inputs generated by adding visually imperceptible perturbations, called adversarial examples, can lure DNNs into working in abnormal ways, which pose potential threats to DNNs based applications, e.g., face recognition \cite{sharif2016accessorize}, autonomous driving \cite{eykholt2018robust}, and malware classification \cite{grosse2016adversarial}. The broad deployment of deep re-ID in security related systems makes it critical to figure out whether such adversarial examples also exist on deep re-ID models. Serious consequences will be brought if deep re-ID systems are proved to be vulnerable to adversarial attacks, for example, a suspect who utilizes this vulnerability can escape from the person search of re-ID based surveillance systems.

To the best of our knowledge, we are the first to investigate robust physical-world attacks on deep re-ID. In this paper, we propose a novel attack algorithm, called advPattern, to generate adversarially transformable patterns across camera views that cause image mismatch in deep re-ID systems. An adversary cannot be correctly matched by deep re-ID models by printing the adversarial pattern on his clothes, like wearing an ``invisible cloak''. We present two different kinds of attacks in this paper: Evading Attack and Impersonation Attack. The former can be viewed as an untargeted attack that the adversary attempts to fool re-ID systems into matching him as an arbitrary person except himself. The latter is a targeted attack which goes further than Evading Attack: the adversary seeks to lure re-ID systems into mismatching himself as a target person. Figure \ref{fig:example_advPattern} gives an illustration of Impersonation Attack on deep re-ID models.

The main challenge with generating adversarial patterns is that \textit{how to cause deep re-ID systems to fail to correctly match the adversary's images across camera views with the same pattern on clothes}. Furthermore, the adversary might be captured by re-ID systems in any position, \textit{ but the adversarial pattern generated specifically for one shooting position is difficult to remain effective in other varying positions}. In addition, other challenges with physically realizing attacks also exist: (1) \emph{How to allow cameras to perceive the adversarial patterns but avoid
arousing suspicion of human supervisors}? (2) \emph{How to make the generated adversarial patterns survive in various physical conditions, such as printing process, dynamic environments and shooting distortion of cameras}?

\begin{figure}[!t]
  \centering
      \includegraphics[width=0.7\columnwidth]{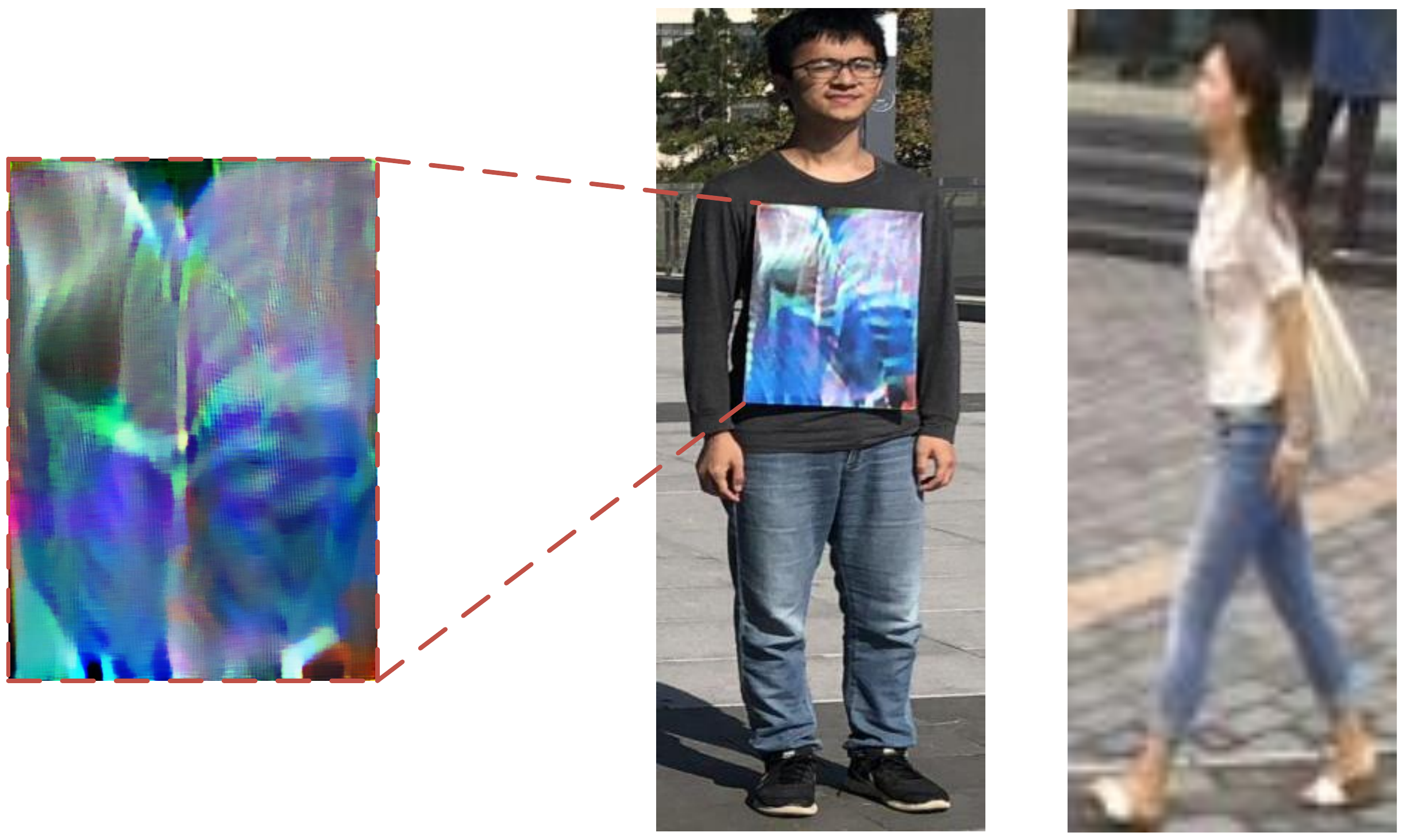}
  \caption{An example of Impersonation Attack in physical world. Left: the digital adversarial pattern; Middle: the adversary wearing a clothes   with the physical adversarial pattern; Right: The target person randomly chosen from Market1501 dataset.}
  \label{fig:example_impersonate}
  \vspace{-4mm}
\end{figure}

To address these challenges, we propose advPattern that formulates the problem of generating adversarial patterns against deep re-ID models as an optimization problem of minimizing the similarity scores of the adversary's images across camera views. The key idea behind advPattern is to amplify the difference of person images across camera views in the process of extracting features of images by re-ID models. To achieve the scalability of adversarial patterns, we approximate the distribution of viewing transformation with a multi-position sampling strategy.
We further improve adversarial patterns' robustness by modeling physical dynamics (e.g., weather changes, shooting distortion), to ensure them survive in physical-world scenario. Figure \ref{fig:example_impersonate} shows an example of our physical-world attacks on deep re-ID systems.

To demonstrate the effectiveness of advPattern, we first establish a new dataset, PRCS, which consists of 10,800 cropped images of 30 identities, and then evaluate the attack ability of adversarial patterns on two deep re-ID models using the PRCS dataset and the publicly available Market1501 dataset. We show that our adversarially transformable patterns generated by advPattern achieve high success rates under both Evading Attack and Impersonation Attack: the rank-1 accuracy of re-ID models for matching the adversary decreases from 87.9\% to 27.1\% under Evading Attack, meanwhile the adversary can impersonate as a target person with 47.1\% rank-1 accuracy and 67.9\% mAP under Impersonation Attack. The results demonstrate that deep re-ID models are indeed vulnerable to our proposed physical-world attacks.

In summary, our main contributions are three-fold:
\begin{itemize}
 \item To the best of our knowledge, we are the first to implement physical-world attacks on deep re-ID systems, and reveal the vulnerability of deep re-ID modes.
 \item We design two different attacks, Evading Attack and Impersonation Attack, and propose a novel attack algorithm advPattern for generating adversarially transformable 
      patterns, to realize adversary mismatch and target person impersonation, respectively.
 \item We evaluate our attacks with two state-of-the-art deep re-ID models and demonstrate the effectiveness of the generated patterns to attack deep re-ID in both digital domain and physical world with high success rate.
\end{itemize}

The remainder of this paper is organized as follows: we review some related works in Section \ref{sec:related} and introduce the system model in Section \ref{sec:system}. In Section \ref{sec:attack}, we present the attack methods for implementing physical-world attacks on deep re-ID models. We evaluate the proposed attacks and demonstrate the effectiveness of our generated patterns in Section \ref{sec:experiments} and conclude with Section \ref{sec:conclusion}.

\section{Related Work}\label{sec:related}

{\bf Deep Re-ID Models.} With the development of deep learning and increasing volumes of available datasets, deep re-ID models have been adopted to automatically learn better feature representation and similarity metric \cite{ahmed2015improved,chen2016deep,chen2017multi,cheng2016person,ding2015deep,li2014deepreid,wang2016joint,xiao2016learning,yi2014deep}, achieving state-of-the art performance. Some methods treat re-ID as a classification issue: Li et al.\ \cite{li2014deepreid} proposed a filter pairing neural network to automatically learn feature representation. Yi et al.\ \cite{yi2014deep} used a siamese deep neural network to solve the re-ID problem. Ahmed et al.\ \cite{ahmed2015improved} added a different matching layer to improve original deep architectures. Xiao et al.\ \cite{xiao2016learning} utilized multi-class classification loss to train model with data from multiple domains. Other approaches solve re-ID as a ranking task: Ding et al.\ \cite{ding2015deep} trained the network with the proposed triplet loss. Cheng et al.\ \cite{cheng2016person} introduced a new term to the original triplet loss to improve model performance. Besides, two recent works \cite{chen2017multi,wang2016joint} considered two tasks simultaneously and built networks to jointly learn representation from classification loss and ranking loss during training.

{\bf Adversarial Examples.} Szegedy et al.\ \cite{szegedy2013intriguing} discovered that neural networks are vulnerable to adversarial examples. Given a DNNs based classifier $f(\cdot)$ and an input $x$ with ground truth label $y$, an adversarial example $x'$ is generated by adding small perturbations to $x$ such that the classifier makes a wrong prediction, as $f(x')\neq y$, or $f(x')=y^{*}$ for a specific target $y^{*}\neq y$. Existing attack methods generate adversarial examples either by one-step methods, like the Fast Gradient Sign Method (FGSM) \cite{goodfellow6572explaining}, or by solving optimization problems iteratively, such as L-BFGS \cite{szegedy2013intriguing}, Basic Iterative Methods(BIM) \cite{kurakin2016adversarial}, DeepFool \cite{moosavi2016deepfool}, and Carlini and Wagner Attacks(C\&W) \cite{carlini2017towards}. Kurakin et al.\ \cite{kurakin2016adversarial} explored adversarial attack in physical world by printing adversarial examples on paper to cause misclassification when photographed by cellphone camera. Sharif et al.\ \cite{sharif2016accessorize} designed eyeglass frame by printing adversarial perturbations on it to attack face recognition systems. Evtimov et al.\  \cite{eykholt2018robust} created adversarial road sign to attack road sign classifiers under different physical conditions. Athalye et al.\ \cite{athalye2017synthesizing} constructed physical 3D-printed adversarial objects to fool a classifier when photographed over a variety of viewpoints.

In this paper, to the best of our knowledge, we are the first to investigate physical-world attacks on deep re-ID models, which differs from prior works targeting on classifiers as follows: (1) Existing works on classification task failed to generate transformable patterns across camera views against image retrieval problems. (2) Attacking re-ID systems in physical world faces more complex physical conditions, for instance, adversarial patterns should survive in printing process, dynamic environments and shooting distortion under any camera views. These differences make it impossible to directly apply existing physical realizable methods on classifiers to attack re-ID models.

\section{System Model}\label{sec:system}

In this section, we first present the threat model and then introduce the our design objectives.

\subsection{Threat Model}\label{sec:threat}

Our work focuses on physically realizable attacks against DNNs based re-ID systems, which capture pedestrians in real-time and automatically search a person of interest across non-overlapping cameras. By comparing the extracted features of a probe (the queried image) with features from a set of continuously updated gallery images collected from other cameras in real time, a re-ID system outputs images from the gallery which are considered to be the most similar to the queried image. We choose re-ID system as our target model because of the wild deployment of deep re-ID in security-critical settings, which will throw dangerous threats if successfully implementing physical-world attacks on re-ID models. For instance, a criminal can easily escape from the search of re-ID based surveillance systems by physically deceiving deep re-ID models.

We assume the adversary has \emph{white-box access} to well-trained deep re-ID models, so that he has knowledge of model structure and parameters, and \emph{only} implements attacks on re-ID models in the inference phase. The adversary is not allowed to manipulate either the digital queried image or gallery images gathered from cameras. Moreover, the adversary is not allowed to change his physical appearance during attacking re-ID systems in order to avoid arousing human supervisor's suspicion. These reasonable assumptions make it challenging to realize successfully physical-world attacks on re-ID systems.


Considering that the stored video recorded by cameras will be copied and re-ID models will be applied for person search only when something happens, the adversary has no idea of when he will be treated as the person of interest and which images will be picked for image matching, which means that the queried image and gallery images are completely unknown to the adversary. However, with the white-box access assumption, the adversary is allowed to construct a generating set $X$ by taking images at each different camera view, which can be realized by stealthily placing cameras at the same position of surveillance cameras to capture images before implementing attacks.

\subsection{Design Objectives}\label{sec:objective}

We propose two attack scenarios, Evading Attack and Impersonation Attack, to deceive deep-ID models.

{\bf Evading Attack.} An Evading Attack is an \emph{untargeted attack}: \emph{Re-ID models are fooled to match the adversary as an arbitrary person except himself, which looks like that the adversary wears an ``invisible cloak''}. Formally, a re-ID model ${f_\theta }\left( { \cdot ,\left. \cdot  \right)} \right.$ outputs a similarity score of an image pair, where $\theta$ is the model parameter. Given a probe image $p_{adv}$ of an adversary, and an image $g_{adv}$ belonging to the adversary in the gallery $G_t$ at time $t$, we attempt to find an adversarial pattern $\delta$ attached on the adversary's clothes to fail deep re-ID models in person search by solving the following optimization problem:
\begin{equation}
\max D(\delta ),\;\;\;s.t.\;Rank({f_\theta }\left( {{p_{adv + \delta }},\left. {{g_{adv + \delta }}} \right))} \right. > K
\end{equation}
where $D(\cdot)$ is used to measure the reality of the generated pattern. Unlike previous works aiming at generating visually inconspicuous perturbations, we attempt to generate \emph{visible patterns} for camera sensing, while \emph{making generated patterns indistinguishable from naturally decorative pattern on clothes}. $Rank(\cdot)$ is a sort function which ranks similarity scores of all gallery images with $p_{adv}$ in the decreasing order. An adversarial pattern is successfully crafted only if the image pair $(p_{adv + \delta }, g_{adv + \delta })$ ranks behind the top-$K$ results, which means that the re-ID systems cannot realize cross-camera image match of the adversary.

{\bf Impersonation Attack.} An Impersonation Attack is a \emph{targeted attack} which can be viewed as an extension of Evading Attack: The adversary attempts to \emph{deceive re-ID models into mismatching himself as a target person}. Given our target's image ${I_t}$, we formulate Impersonation Attack as the following optimization problem:
\begin{equation}
\max D(\delta ), \;\; s.t.\left\{\begin{array}{l}
\hspace{-2mm} Rank ({f_\theta }\left( {{p_{adv + \delta }},\left. {{g_{adv + \delta }}} \right))} \right. > K\\
\hspace{-2mm} Rank ({f_\theta }\left( {{p_{adv + \delta }},\left. {{I_t}} \right))} \right. < K
\end{array} \right. \hspace{-2mm}
\end{equation}
we can see that, besides the evading constraint, the optimization problem for an Impersonation Attack includes another constraint that the image pair $(p_{adv + \delta }, {I_t})$ should be within the top-K results, which implies that the adversary successfully induces the re-ID systems into matching him to the target person.

Since the adversary has no knowledge of the queried image and the gallery, it is impossible for the adversary to solve the above optimization problems. In the following section, we will present the solution that approximately solve the above optimization problems.


\section{Adversarial Pattern Generation}\label{sec:attack}
\begin{figure}[!t]
  \centering
      \includegraphics[width=0.9\columnwidth]{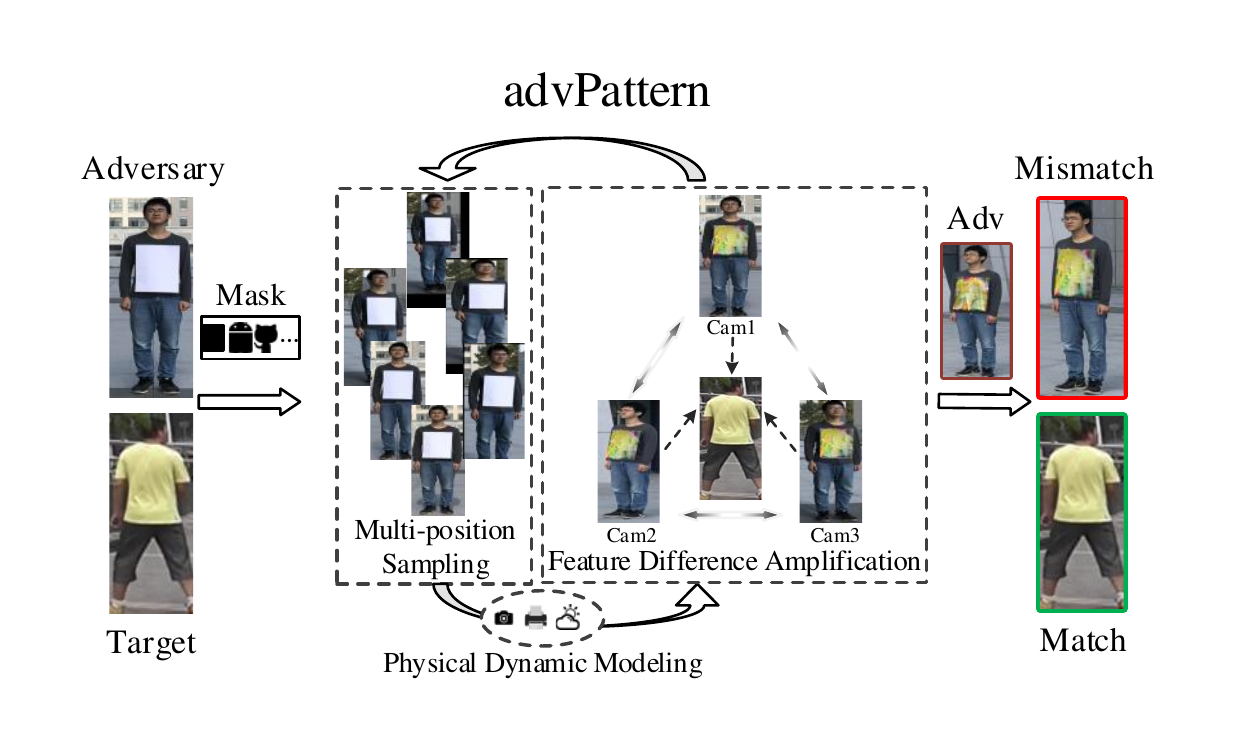}
  \caption{Overview of the attack pipeline.}
  \label{fig3}
  \vspace{-4mm}
\end{figure}
In this section, we present a novel attack algorithm, called advPattern, to generate adversarial patterns for attacking deep re-ID systems in real-world. Figure \ref{fig3} shows an overview of the pipeline to implement an Impersonation Attack in physical world. Specifically, we first generate transformable patterns across camera views for attacking the image retrieval problem as described in Section \ref{sec:transformable}. To implement position-irrelevant and physical-world attacks, we further improve the scalability and robustness of adversarial patterns in Section \ref{sec:scalable} and Section \ref{sec:robust}.

\subsection{Transformable Patterns across Camera Views}\label{sec:transformable}

Existing works \cite{cheng2016person,zhong2018camera} found that there exists a common image style within a certain camera view, while dramatic variations across different camera views. To ensure that the same pattern can cause cross-camera image mismatch in deep re-ID models, we propose an adversarial pattern generation algorithm to generate transformable  patterns that amplify the distinction of the adversary's images across camera views in the process of extracting features of images by re-ID models.

For the {\bf Evading Attack}, given the generating set $X =(x_{1},x_{2},...,x_{m})$ constructed by the adversary, which consists of the adversary's images captured from $m$ different camera views. For each image $x_i$ from $X$, we compute the adversarial image ${x_i}^\prime = o({x_i},{T_i}(\delta))$.  $o({x_i},{T_i}(\delta))$ denotes overlaying the corresponding areas of $x_i$ after transformation $T_{i}(\cdot)$ with the generated pattern $\delta$. Here $T_{i}(\delta)$ is a perspective transformation operation of the generated pattern $\delta$, which ensures the generated pattern to be in accordance with transformation on person images across camera views. We generate the transformable adversarial pattern $\delta$ by solving the following optimization problem:
\begin{equation}\label{eq:3}
\mathop {\arg \min }\limits_\delta \sum\limits_{{\rm{i}} = 1}^m {\sum\limits_{j = 1}^m {{f_\theta }({x_i}^\prime ,{x_j}^\prime )} } , \;\;s.t.\;\;\;i \ne j
\end{equation}
We iteratively minimize the similarity scores of images of an adversary from different cameras to gradually pull farther extracted features of the adversary's images from different cameras by the generated adversarial pattern.

For the {\bf Impersonation Attack}, given a target person's image ${I_t}$, we optimize the following problem:
\begin{equation}
\begin{split}
\mathop {\arg \min }\limits_\delta &\sum\limits_{{\rm{i}} = 1}^m {\sum\limits_{j = 1}^m {{f_\theta }({x_i}^\prime ,{x_j}^\prime )} }  \\
&- \alpha ({f_\theta }({x_i}^\prime ,{I_{\rm{t}}}) + {f_\theta }({x_j}^\prime ,{I_{\rm{t}}})) , \;\;s.t.\;\;\;i \ne j
\end{split}\label{eq:4}
\end{equation}
where $\alpha$ controls the strength of different objective terms. By adding the second term in Eq. \ref{eq:4}, we additionally maximize similarity scores of the adversary's images with the target person's image to generate a more powerful adversarial pattern to pull closer the extracted features of the adversary's images and the target person's image.

\subsection{Scalable Patterns in Varying Positions}\label{sec:scalable}

The adversarial patterns should be capable of implementing successful attacks at any position, which means our attacks should be position-irrelevant. To realize this objective, we further improve the scalability of the adversarial pattern in terms of varying positions.


Since we cannot capture the exact distribution of viewing transformation, we augment the volume of the generating set with a multi-position sampling strategy to approximate the distribution of images for generating scalable adversarial patterns. The augmented generating set $X^{C}$ for an adversary is built by collecting the adversary's images with various distances and angles from each camera view, and synthesized instances generated by image transformation such as translation and scaling on original collected images.

\begin{figure}[!t]
  \centering
      \includegraphics[width=0.6\columnwidth]{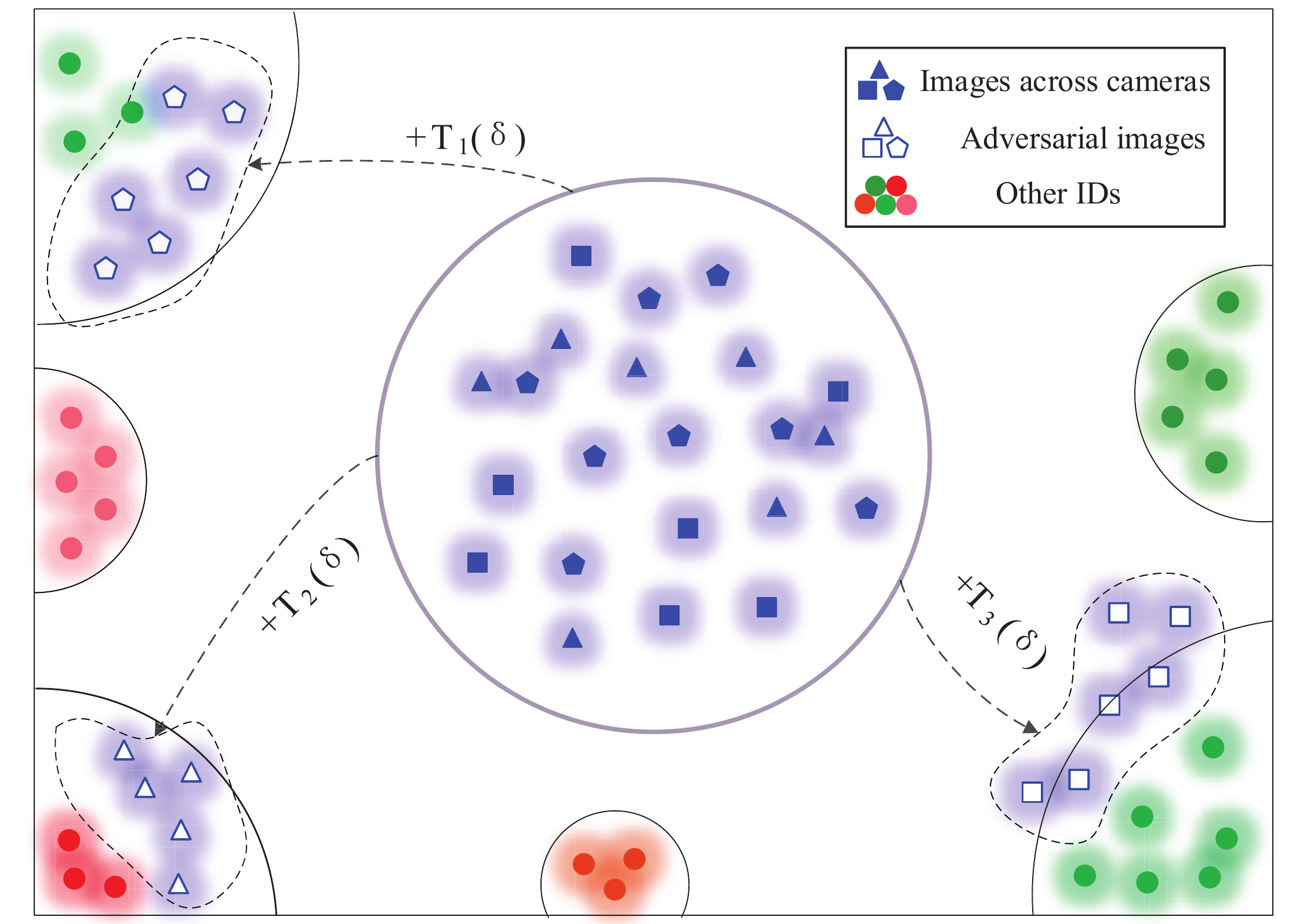}
  \caption{The illustration of how scalable adversarial patterns work. By adding the generated adversarial pattern, the adversarial images from the same camera view are clustered together in the feature space. Meanwhile, the distance of adversarial images from different cameras becomes farther.}
    \label{fig4}
  \vspace{-4mm}
\end{figure}

For the {\bf Evading Attack}, given a triplet ${tri_k} = < x_k^o,x_k^ + ,x_k^ - >$ from $X^{C}$, where $x_k^o$ and $x_k^ +$ are person images from the same camera, while $x_k^ -$ is the person image from a different camera, for each image $x_k$ from ${tri_k}$, we compute the adversarial image ${x_k}^\prime$ as $o({x_k},{T_k}(\delta))$. We randomly chose a triplet at each iteration for solving the following optimization problem:
\begin{equation}\label{eq:3}
\mathop {\arg \min }\limits_\delta {\mathbb{E}_{_{{tri_k} \sim {X^C}}}}{f_\theta }((x_k^o)',(x_k^ - )') - \beta {f_\theta }((x_k^o)',(x_k^ + )')
\end{equation}
where $\beta$ is a hyperparameter that balances different objectives during optimization. The objective of Eq.(\ref{eq:3}) is to minimize the similarity scores of $x_k^o$ with $x_k^ -$ to discriminate person images across camera views, while maximizing similarity scores of $x_k^o$ with $x_k^ +$ to preserve the similarity of person images from the same camera view. During optimization the generated pattern learns the scalability from the augmented generating set $X^{C}$ to pull closer the extracted features of person images from the same camera, while pushing features from different cameras farther, as shown in Figure \ref{fig4}.

For the {\bf Impersonation Attack}, given an image set ${I^t}$ of the target person, and a quadruplet ${quad_k} = < x_k^o,x_k^ + ,x_k^ - ,{t_k}>$ consisting of a triplet ${tri_k}$ and a person image $t_k$ from ${I^t}$, we randomly choose a quadruplet at each iteration, and iteratively solve the following optimization problem:
\begin{equation}
\begin{split}
\mathop {\arg \max }\limits_\delta &{\mathbb{E}_{_{{quad_k} \sim \{ {X^C},{I^t}\} }}}{f_\theta }((x_k^o)',{t_k}) \\
&+ {\lambda _1}{f_\theta }((x_k^o)',(x_k^ + )') - {\lambda _2}{f_\theta }((x_k^o)',(x_k^ - )')
\end{split}
\end{equation}
where $\lambda _1$ and $\lambda _2$ are hyperparameters that control the strength of different objectives. We add an additional objective that maximizes the similarity score of $x_k^o$ with $t_k$ to pull closer the extracted features of the adversary's images to the features of the target person's images.

\subsection{Robust Patterns for Physically Realizable Attack}\label{sec:robust}

Our goal is to implement physically realizable attacks on deep re-ID systems by generating physically robust patterns on adversaries' clothes. To ensure adversarial patterns to be perceived by cameras, we generate large magnitude of patterns with no constraints over them during optimization. However, introducing conspicuous patterns will in turn make adversaries be attractive  and arouse suspicion of human supervisors.

To tackle this problem, we design unobtrusive adversarial patterns which are visible but difficult for humans to distinguish them from the decorative patterns on clothes. To be specific, we choose a mask $M_x$ to project the generated pattern to a shape that looks like a decorative pattern on clothes (e.g., commonplace logos or creative graffiti). In addition, to generate smooth and consistent patches in our pattern, in other words, colors change only gradually within patches, we follow Sharif et al.\ \cite{sharif2016accessorize} that adds total variation ($TV$) \cite{mahendran2015understanding} into the objective function:
\begin{equation}
TV(\delta) = \sum\limits_{p,q} {{{({{({\delta _{p,q}} - {\delta _{p + 1,q}})}^2} + {{({\delta _{p,q}} - {\delta _{p,q + 1}})}^2})}^{\frac{1}{2}}}}
\end{equation}
where $\delta_{p,q}$ is a pixel value of the pattern $\delta$ at coordinates $(p,q)$, and $TV(\delta)$ is high when there are large variations in the values of adjacent pixels, and low otherwise. By minimizing $TV(\delta)$, the values of adjacent pixels are encouraged to be closer to each other to improve the smoothness of the generated pattern.

Implementing physical-world attacks on deep re-ID systems requires adversarial patterns to survive in various environmental conditions. To deal with this problem, we design a degradation function $\varphi(\cdot)$ that randomly changes the brightness or blurs the adversary's images from the augmented generating set $X^{C}$. During optimization we replace $x_{i}$ with the degraded image $\varphi(x_{i})$  to improve the robustness of our generated pattern against physical dynamics and shooting distortion. Recently, the non-printability score (NPS) was utilized in \cite{eykholt2018robust,sharif2016accessorize} to account for printing error. We introduce NPS into our objective function but find it hard to balance NPS term with other objectives. Alternatively, we constrain the search space of the generated pattern $\delta$ in a narrower interval $\mathbb{P}$ to avoid unprintable colors (e.g., high brightness and high saturation). Thus, for each image $x_k$ from ${tri_k}$, we use $o({x_k},{T_k}(M_x\cdot\delta))$ to compute the adversarial images ${x_k}^\prime$, and generate robust adversarial patterns to implement physical-world attack as solving the following optimization problem:
\begin{equation}
\begin{split}
&\mathop {\arg \min }\limits_\delta {\mathbb{E}_{_{{tri_k} \sim {X^C}}}} {f_\theta }(\varphi (x_k^o)',\varphi (x_k^ - )') \\
&- \beta {f_\theta }(\varphi (x_k^o)',\varphi (x_k^ + )') + \kappa  \cdot TV(\delta), \;\;s.t.\;\;\;\delta  \in \mathbb{P}
\end{split}
\end{equation}
where $\lambda$ and $\kappa$ are hyperparameters that control the strength of different objectives. Similarly, the formulation of the Impersonation Attack is analogous to that of the Evading Attack, which is as follows:
\begin{equation}
\begin{split}
&\mathop {\arg \max }\limits_\delta {\mathbb{E}_{_{{quad_k} \sim \{ {X^C},{I^t}\} }}}{f_\theta }(\varphi(x_k^o)',{t_k}) \\
&+ {\lambda _1}{f_\theta }(\varphi(x_k^o)',\varphi(x_k^ + )') - {\lambda _2}{f_\theta }(\varphi(x_k^o)',\varphi(x_k^ - )') \\
&+ \kappa  \cdot TV(\delta ), \;\;s.t.\;\;\;\delta  \in \mathbb{P}
\end{split}
\end{equation}
Finally, we print the generated pattern over the the adversary's clothes to deceive re-ID into mismatching him as an arbitrary person or a target person.

\section{Experiments}\label{sec:experiments}

In this section, we first introduce the datasets and the target deep re-ID models used for evaluation in Section \ref{sec:data}. We evaluate the proposed advPattern for attacking deep re-ID tools both under digital environment (Section \ref{sec:digital}) and in physical world (Section \ref{sec:physical}). We finally discuss the implications and limitations of advPattern in Section \ref{sec:discuss}.

\begin{figure}[!t]
  \centering
      \includegraphics[width=0.4\columnwidth]{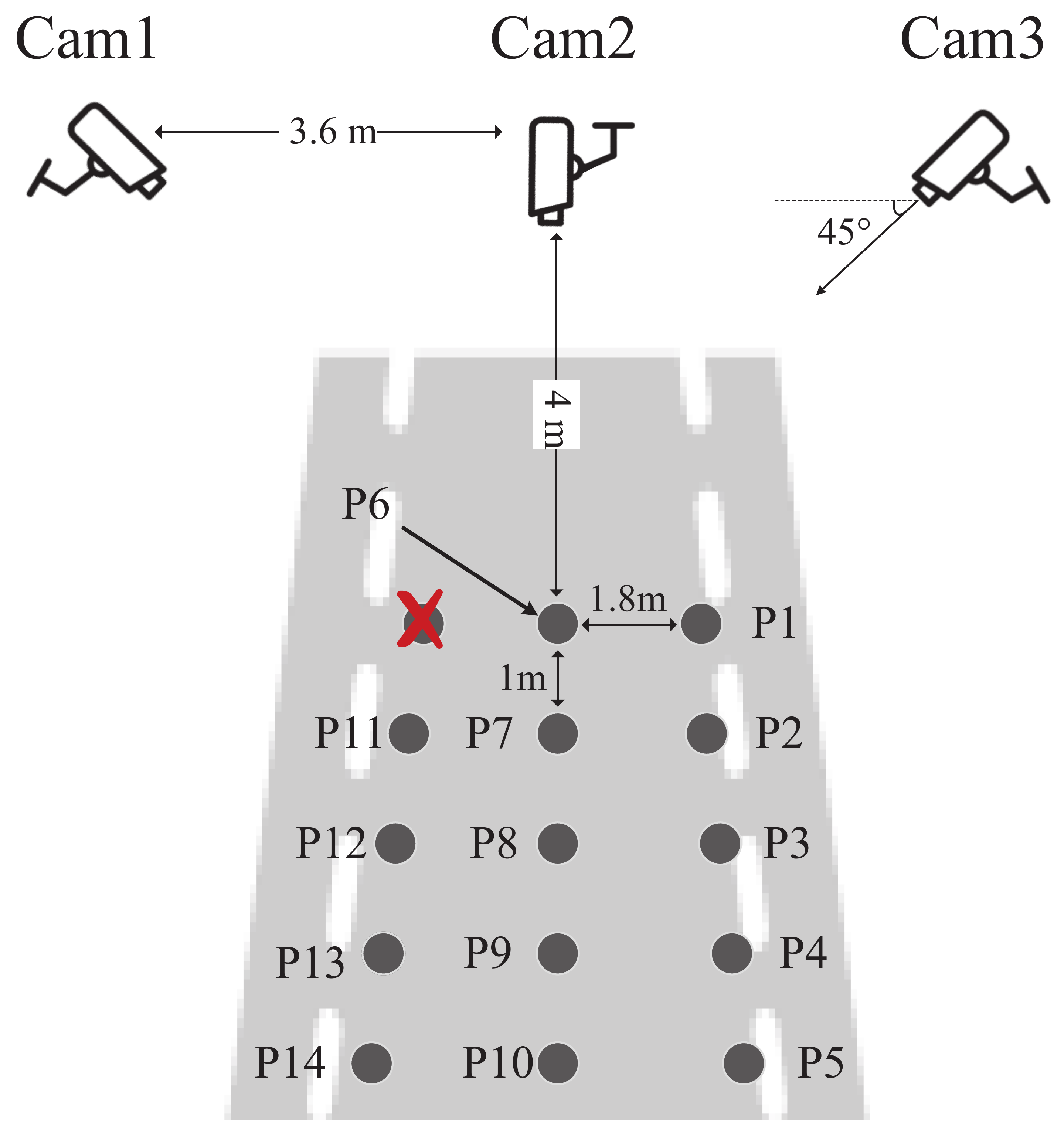}
  \caption{The scene setting of physical-world tests. We choose 14 testing points under each camera which vary in distances and angles.}
  \label{fig5}
  \vspace{-4mm}
\end{figure}

\subsection{Datasets and re-ID Models}\label{sec:data}

{\bf Market1501 Dataset.} Market1501 contains 32,668 annotated bounding boxes of 1501 identities, which is divided into two non-overlapping subsets: the training dataset contains 12,936 cropped images of 751 identities, while the testing set contains 19,732 cropped images of 750 identities.

{\bf PRCS Dataset.} We built a Person Re-identification in Campus Streets (PRCS) dataset for evaluating the attack method. PRCS contains 10,800 cropped images of 30 identities. During dataset collection, three cameras were deployed to capture pedestrians in different campus streets. Each identity in PRCS was captured by both three cameras and has at least 100 cropped images per camera. We chose 30 images of each identity per camera to construct the testing dataset for evaluating the performance of the trained re-ID models and our attack method.

\begin{table}[!t]
{\footnotesize
\tabcolsep 5.2pt
\caption{Re-ID performance of model A and model B on Market1501 and PRCS datasets. (ss = Similarity Score)}
\vspace{5pt} \label{tab1}
\begin{tabular}{ccllllp{2mm}}
\toprule
Model                                   & Dataset                          & \multicolumn{1}{c}{rank-1} & \multicolumn{1}{c}{rank-5} & \multicolumn{1}{c}{rank-10} & \multicolumn{1}{c}{mAP}  & \multicolumn{1}{c}{ss}    \\ \hline

\multicolumn{1}{c|}{\multirow{2}{*}{A}} & \multicolumn{1}{c|}{Market-1501} & \multicolumn{1}{c}{77.8\%}   & \multicolumn{1}{c}{90.2\%}   & \multicolumn{1}{c}{93.5\%}    & \multicolumn{1}{c}{62.7\%} & \multicolumn{1}{c}{0.796} \\ \cline{2-2}
\multicolumn{1}{c|}{}                   & \multicolumn{1}{c|}{PRCS}        & \multicolumn{1}{c}{87.9\%}   & \multicolumn{1}{c}{93.4\%}   & \multicolumn{1}{c}{100.0\%}   & \multicolumn{1}{c}{78.6\%} & \multicolumn{1}{c}{0.876} \\ \hline
\specialrule{0em}{-2pt}{-5pt}\\ \hline
\multicolumn{1}{c|}{\multirow{2}{*}{B}} & \multicolumn{1}{c|}{Market-1501} & \multicolumn{1}{c}{74.5\%}   & \multicolumn{1}{c}{89.0\%}   & \multicolumn{1}{c}{92.8\%}    &                          \multicolumn{1}{c}{57.3\%} & \multicolumn{1}{c}{0.732} \\ \cline{2-2}
\multicolumn{1}{c|}{}                   & \multicolumn{1}{c|}{PRCS}        & \multicolumn{1}{c}{84.7\%}   & \multicolumn{1}{c}{95.4\%}   & \multicolumn{1}{c}{99.0\%}     &                          \multicolumn{1}{c}{77.2\%} & \multicolumn{1}{c}{0.857} \\ \bottomrule
\end{tabular}
}
\end{table}

\begin{table}[!t]
{\footnotesize
\tabcolsep 6.5pt
\caption{Digital-environment attack results on model A and model B under Evading Attack (GS = Generating Set, TS = Testing Set).}
\vspace{5pt} \label{tab2}
\begin{tabular}{cclllll}
\toprule
Model                                   & Dataset                            & \multicolumn{1}{c}{rank-1} & \multicolumn{1}{c}{rank-5} & \multicolumn{1}{c}{rank-10} & \multicolumn{1}{c}{mAP} & \multicolumn{1}{c}{ss}    \\ \hline
\multicolumn{1}{c|}{\multirow{2}{*}{A}} & \multicolumn{1}{c|}{GS}  & \multicolumn{1}{c}{0.0\%}    & \multicolumn{1}{c}{0.0\%}    & \multicolumn{1}{c}{0.0\%}     & \multicolumn{1}{c}{4.4\%} & \multicolumn{1}{c}{0.394} \\ \cline{2-2}
\multicolumn{1}{c|}{}                   & \multicolumn{1}{c|}{TS}   & \multicolumn{1}{c}{4.2\%}    & \multicolumn{1}{c}{8.3\%}    & \multicolumn{1}{c}{16.7\%}    & \multicolumn{1}{c}{7.3\%} & \multicolumn{1}{c}{0.479} \\ \hline
\specialrule{0em}{-2pt}{-5pt}\\ \hline
\multicolumn{1}{c|}{\multirow{2}{*}{B}} & \multicolumn{1}{c|}{GS} & \multicolumn{1}{c}{0.0\%}    & \multicolumn{1}{c}{0.0\%}    & \multicolumn{1}{c}{0.0\%}     & \multicolumn{1}{c}{4.5\%} & \multicolumn{1}{c}{0.422} \\ \cline{2-2}
\multicolumn{1}{c|}{}                   & \multicolumn{1}{c|}{TS}   & \multicolumn{1}{c}{10.4\%}    & \multicolumn{1}{c}{13.3\%}    & \multicolumn{1}{c}{16.7\%}    & \multicolumn{1}{c}{16.3\%} & \multicolumn{1}{c}{0.508} \\ \bottomrule
\end{tabular}
\label{tab:digital}
}
\vspace{-4mm}
\end{table}

{\bf Target Re-ID Models.} We evaluated the proposed attack method on two different types of deep re-ID models:  model A is a siamese network proposed by Zheng et al. \cite{zheng2017discriminatively}, which is trained by combining verification loss and identification loss; model B utilizes a classification model to learn the discriminative embeddings of identities as introduced in \cite{zheng2016person}. The reason why we choose the two models as target models is that classification networks and siamese networks are widely used in the re-ID community. The effectiveness of our attacks on the two models can imply the effectiveness on other models. Both of the two models achieve the state-of-the-art performance (i.e., rank-$k$ accuracy and mAP) on Market1501 dataset, and also work well on PRCS dataset. The results are given in Table \ref{tab1}.

We use the ADAM optimizer to generate adversarial patterns with the following parameters setting: learning rate $=0.01$, $\beta_1 = 0.9$, $\beta_2 = 0.999$. We set the maximum number of iterations to 700.

\begin{table*}[!ht]
\centering
{\footnotesize
\caption{Digital-environment attack results on target models under Impersonation Attack. The performance of matching the adversary as the target person (Target) and himself (Adversary) are both given. (GS = Generating Set, TS = Testing Set).}
\vspace{5pt}
\begin{tabular}{c|c|>{\columncolor[gray]{0.8}}c>{\columncolor[gray]{0.8}}c>{\columncolor[gray]{0.8}}c>{\columncolor[gray]{0.8}}c>{\columncolor[gray]{0.8}}c|
>{\columncolor[gray]{0.8}}c>{\columncolor[gray]{0.8}}c>{\columncolor[gray]{0.8}}c>{\columncolor[gray]{0.8}}c>{\columncolor[gray]{0.8}}c}
\toprule
\rowcolor{white}
& &   \multicolumn{5}{c|}{PRCS} & \multicolumn{5}{c}{Market1501} \\
\rowcolor{white} \multirow{-2}{*}{Model} & \multirow{-2}{*}{Matched person} & rank-1 & rank-5 & rank-10 & mAP  & ss    & rank-1 & rank-5 & rank-10 & mAP  & ss
\\ \hline

&  Target(GS)   & 86.7\%   & 97.2\%   & 100.0\%   & 89.7\% & 0.848 & 68.0\%   & 83.8\%   & 89.8\%    & 70.7\% & 0.814      \\
\rowcolor{white} \multirow{-2}{*}{A} & Adversary(GS)                   & 5.50\%    & 5.70\%    & 8.30\%     & 8.10\%  & 0.524 & 4.10\%    & 4.10\%    & 10.4\%    & 9.20\%  & 0.565                   \\ \hline

&  Target(GS)   & 92.8\%   & 97.7\%   & 100.0\%   & 91.8\% & 0.858 & 94.0\%   & 98.1\%   & 100.0\%   & 60.7\% & 0.775      \\
\rowcolor{white} \multirow{-2}{*}{B} & Adversary(GS)                   & 2.50\%    & 5.80\%    & 8.40\%     & 5.50\%  & 0.486 & 1.80\%    & 6.90\%    & 8.70\%     & 10.9\% & 0.606                   \\ \hline

&  Target(TS)   & 74.4\%   & 83.3\%   & 91.7\%    & 81.5\% & 0.824\% & 55.8\%   & 70.8\%   & 79.2\%    & 45.1\% & 0.803      \\
\rowcolor{white} \multirow{-2}{*}{A} & Adversary(TS)                   & 19.4\%   & 22.2\%  & 38.9\%    & 18.4\% & 0.633 & 14.1\%   & 17.1\%   & 27.5\%    & 12.9\% & 0.638                   \\ \hline

&  Target(TS)   & 78.4\%   & 83.3\%   & 91.7\%    & 88.2\% & 0.812 & 68.7\%   & 79.2\%   & 91.6\%    & 51.9\% & 0.749      \\
\rowcolor{white} \multirow{-2}{*}{B} & Adversary(TS)                   & 16.7\%   & 18.9\%   & 41.7\%    & 24.8\% & 0.652 & 28.4\%   & 34.7\%   & 50.4\%    & 31.3\% & 0.659                    \\ \bottomrule
\end{tabular}
\label{tab:digital_impersonate}
}
\end{table*}

\begin{table*}[!ht]
\centering
{\footnotesize
\caption{Physical-world attack results on target models at varying distances and angles. Distance\&Angle of each points with camera 3 are given. $\Delta$rank-1, $\Delta$mAP and $\Delta$ss indicate the drop of target models' performance due to adversarial patterns.}
\vspace{5pt} \label{tab4}
\begin{tabular}{cccccccccc}
\toprule
                                  & \multicolumn{6}{c}{Evading Attack}                           & \multicolumn{3}{c}{Impersonation Attack}                       \\
\multirow{-2}{*}{Distance\&Angle} & rank-1 & $\Delta$rank-1      & mAP & $\Delta$mAP           & ss & $\Delta$ss             & rank-1                     & mAP           & ss             \\ \hline
  P1 (4.39, 24.2)           & 0.00\% & 100\%       & 18.9\% & 69.2\%          & 0.689 & 0.146         & 40.0\%    & 58.4\%    & 0.774          \\
  P2 (5.31, 19.8)           & 20.0\% &80.0\%        & 25.6\%  & 66.5\%       & 0.694 & 0.149        & 80.0\%       & 83.8\%     & 0.746          \\
  P3 (6.26, 16.7)           & 20.0\% & 80.0\%     & 24.6\% & 66.7\%       & 0.687 & 0.163          & 0.0\%        & 33.6\%     & 0.728          \\
  P4 (7.23, 14.4)          & 20.0\% & 80.0\%       & 21.1\% & 68.9\%      & 0.680 & 0.180         & 80.0\%        & 84.9\%     & 0.737          \\
  P5 (8.20, 12.7)          & 20.0\% & 80.0\%      & 23.4\% & 64.2\%      & 0.660 & 0.124        & 20.0\%      & 58.4\%      & 0.748          \\
  P6 (5.38, 42.0)          & 0.00\%  & 100\%     & 18.6\% & 71.2\%     & 0.685 &0.192         & 40.0\%     & 68.0\%        & 0.784          \\
  P7 (6.16, 35.8)          & 40.0\% & 60.0\%       & 27.6\% & 62.7\%     & 0.709 & 0.164        & 40.0\%      & 65.2\%      & 0.761          \\
  P8 (7.00, 31.0)          & 0.00\%  & 100\%      & 15.9\% & 40.8\%     & 0.663 & 0.083        & 80.0\%    & 85.8\%        & 0.762          \\
  P9 (7.87, 27.2)           & 0.0\% & 100\%       & 18.0\% & 71.6\%    & 0.669 & 0.158        & 40.0\%    & 67.1\%          & 0.767          \\
  P10 (8.77, 24.2)         & 40.0\% & 60.0\%      & 40.0\%& 57.5\%     & 0.714 & 0.167        & 40.0\%     & 59.5\%        & 0.761          \\
  P11 (7.36, 47.2)         & 20.0\% & 80.0\%    & 22.6\% & 65.6\%     & 0.695 & 0.147        & 60.0\%      & 72.1\%          & 0.768          \\
  P12 (8.07, 42.0)         & 60.0\% & 40.0\%    & 27.6\% & 47.6\%      & 0.688 & 0.124        & 80.0\%        & 82.3\%         & 0.771          \\
  P13 (8.84, 37.6)        & 80.0\% & 20.0\%     & 47.5\%& 36.8\%      & 0.749 & 0.071        & 40.0\%      & 78.3\%          & 0.789          \\
  P14 (9.65, 34.0)        & 40.0\% & 60.0\%     & 20.9\%& 66.7\%       & 0.699 & 0.148         & 20.0\%  & 52.6\%          & 0.768          \\ \hline
Average                  & \textbf{27.1\%} & \textbf{74.3\%} & \textbf{25.2\%} & \textbf{61.1\%} & \textbf{0.692} & \textbf{0.144} & \textbf{47.1\%}    & \textbf{67.9\%} & \textbf{0.762} \\ \bottomrule
\end{tabular}
}
\vspace{-5mm}
\end{table*}

\subsection{Digital-Environment Tests}\label{sec:digital}
We first evaluate our attack method in digital domain where the adversary's images are directly modified by digital adversarial patterns$\footnote{The code is avaliable at \url{https://github.com/whuAdv/AdvPattern}}$. It is worth noting that attacking in digital domain is actually not a realistic attack, but a necessary evaluation step before successfully implementing real physical-world attacks.

{\bf Experiment Setup.} We first craft the adversarial pattern over a generating set for each adversary, which consists of real images from varying positions, viewing angles and synthesized samples. Then we attach generated adversarial pattern to the adversary's images in digital domain to evaluate the attacking performance on the target re-ID models.

We choose every identity from PRCS as an adversary to attack deep re-ID. In each query, we choose an image from adversarial images as the probe image, and construct a gallery by combining 12 adversarial images from other cameras with images from 29 identities in PRCS, and 750 identities in Market1501. For Impersonation Attack, we take two identities as target for each adversary: one is randomly chosen from Market1501, and another one is chosen from PRCS. We ran 100 queries for each attack.

{\bf Experiment Results.}
Table \ref{tab:digital} shows the attack results on two re-ID models under Evading Attack in digital environment. We can see that the matching probability and mAP drops significantly for both re-ID models, which demonstrate the high success rate of implementing the Evading Attack. The similarity score of the adversary's images decreases to less than 0.5, making it hard for deep re-ID models to correctly match images of the adversary in the large gallery. Note that the attack performance on testing set is close to generating set, e.g., rank-1 accuracy from 4.2\% to 0\% and mAP from 7.3\% to 4.4\%, which demonstrates the scalability of the digital adversarial patterns when implementing attacks with unseen images.

Table \ref{tab:digital_impersonate} shows the attack results to two re-ID models under Impersonation Attack in the digital environment. In PRCS, the average rank-1 accuracy is above 85\% when matching adversarial images from the generating set as a target person, which demonstrates the effectiveness of implementing the targeted attack. The patterns are less effective when targeting an identity from Market1501: the rank-1 accuracy of model A decreases to 68.0\% for the generating set, and 41.7\% for the testing set. We attribute it to the large variations in physical appearance and image styles between two datasets. Though, the high rank-5 accuracy and mAP demonstrate the strong capability of digital patterns to deceive target models. Note that the rank-$k$ accuracy and mAP decrease significantly for matching the adversary's images across cameras, which means that the generated patterns can also cause mismatch across camera views in targeted attacks. Again, that the attack performance on testing set is close to generating set demonstrates the scalability of the adversarial patterns with unseen images.


\begin{figure}[!t]
  \centering
      \includegraphics[width=0.8\columnwidth]{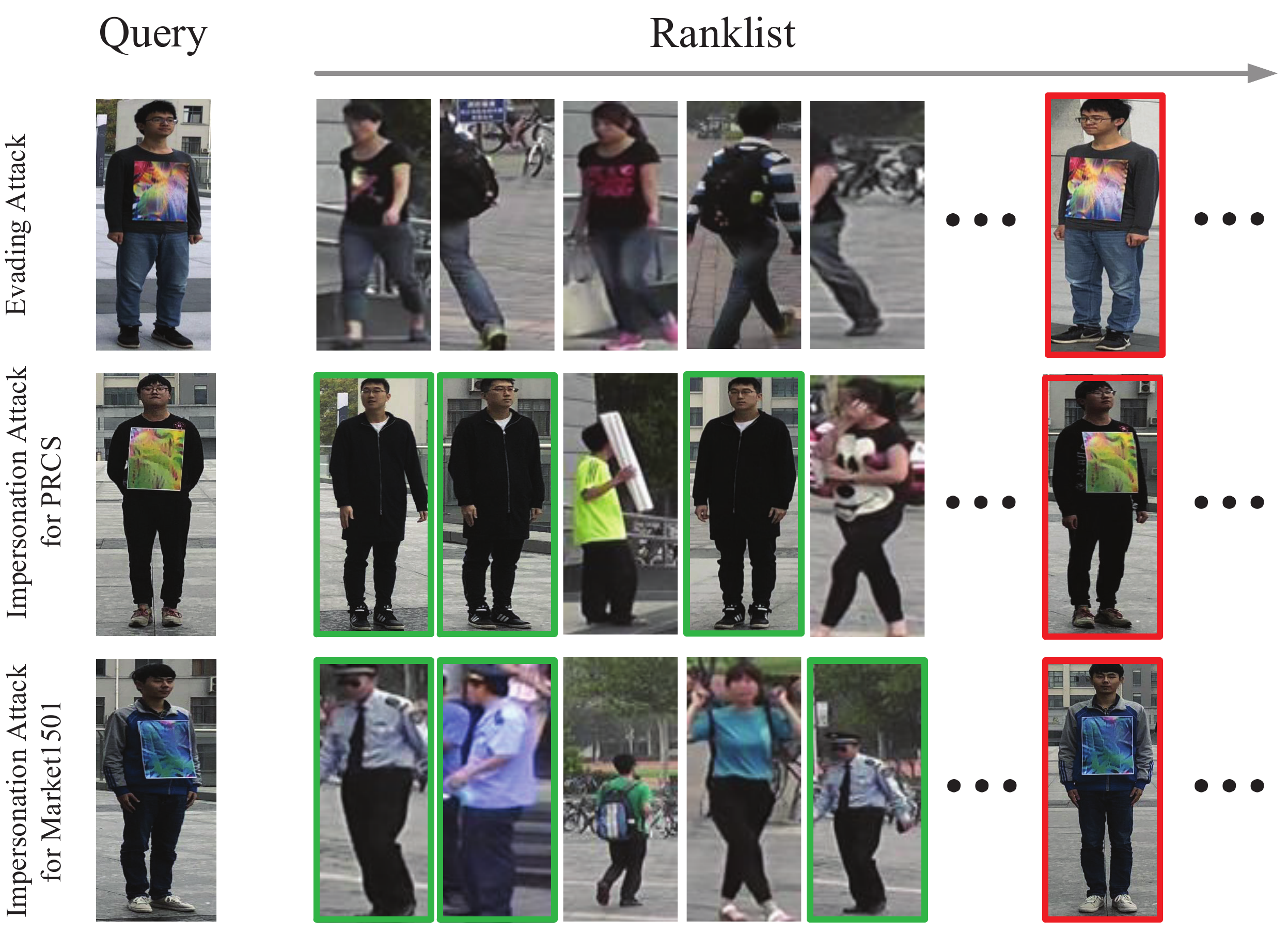}
  \caption{Examples of physically realizable attacks. Top row: an Evading Attack (the adversary: ID1 from PRCS). Middle row: an Impersonation Attack targeting identity from PRCS (the adversary: ID2 in PRCS, the target: ID12 from PRCS). Bottom row: an Impersonation Attack targeting identity from Market1501 (the adversary: ID3 in PRCS, the target: ID728 from Market1501)}
  \label{fig6}
  \vspace{-4mm}
\end{figure}

\subsection{Physical-World Evaluation}\label{sec:physical}

On the basis of digital-environment tests, we further evaluate our attack method in physical world. We print the adversarial pattern and attach it to the adversary's clothes for implementing physical-world attacks.

{\bf Experiment Setup.} The scene setting of physical-world tests is shown in Figure \ref{fig5}, where we take images of the adversary with/without the adversarial pattern in 14 testing points with variations in distances and angles from cameras. These 14 points are sampled with a fix interval in the filed of cameras views. We omit the left-top point in our experiment due to the constraint of shooting conditions. The distance between cameras and the adversary is about from $5m$ to $10m$ for better perceiving the adversarial pattern.

We choose 5 identities as adversaries from PRCS to implement physically realizable attacks. In each query, we randomly choose the adversary's image under a testing point as the probe image,  while adding 12 adversarial images from other cameras into the gallery. Two identities are randomly chosen from Market1501 and PRCS respectively to serve as target person. 100 queries for each testing point are performed. We evaluate physical-world attacks for Evading Attack with model A, while Impersonation Attack with model B.

{\bf Experiment Results.} Table \ref{tab4} shows the physical-world attack results of 14 different testing positions with varying distances and angles from cameras. Note that $\Delta$rank-1 denotes the drop of match probability due to adversarial patterns. Similar meanings happen to $\Delta$mAP and $\Delta$ss. For the Evading Attack, we can see that it significantly decreases the match probability to the adversary with the crafted adversarial pattern. The average $\Delta$rank-1 and $\Delta$mAP are 62.2\% and 61.1\%. The average of the rank-1 accuracy and mAP are 47.1\% and 67.9\% under Impersonation Attack, respectively. The results demonstrates the effectiveness of adversarial patterns to implement physical-world attacks in varying positions with considerable success rate.

For Evading Attack, the average rank-1 accuracy drops to 11.1\% in 9 of 14 positions, which demonstrates that the generated adversarial patterns can physically attack deep re-ID systems with high success rate. Note that adversarial patterns are less effective in some testing points, e.g., P12 and P13. We attribute it to the larger angles and farther distance between these points and cameras, which makes it more difficult for cameras to perceive the patterns. For Impersonation Attack, The rank-1 accuracy for matching the adversary as the target person is 56.4\% in 11 of 14 positions , which is close to the result of digital patterns targeting on Market1501. The high mAP and similarity scores when matching the adversary as the targeted person demonstrate the effectiveness of adversarial patterns to implement targeted attack in physical world. Still, there exists few points (P3, P5, P14) where the adversary has trouble to implement successful attack with adversarial patterns. Figure \ref{fig6} shows examples of physical-world attacks on deep re-ID systems.

\subsection{Discussion} \label{sec:discuss}

{\bf Black Box Attacks.} In this paper, we start with the white-box assumption to investigate the vulnerability of deep re-IDs models. Nevertheless, it would be more meaningful if we can realize adversarial patterns with black-box setting. Prior works \cite{liu2016delving,papernot2016transferability} demonstrated successful attacks without any knowledge of model's internals by utilizing the transferability of adversarial examples. We will leave black-box attacks as our future work.

{\bf AdvPattern vs. Other Approaches.} AdvPattern allows the adversary to deceive deep re-ID systems without any digital modifications of person images or any physical appearance change. Although there are simpler ways to attack re-ID systems, e.g., directly object removal in digital domain, or changing physical appearance in different camera view, we argue that our adversarial pattern is the most reasonable method because: (1) for object removal methods, it is unrealistic to control the queried image and gallery images; (2) changing physical appearance makes adversaries attractive to human supervisors.

\section{Conclusion}\label{sec:conclusion}
This paper designed Evading Attack and Impersonation Attack for deep re-ID systems, and proposed advPattern for generating adversarially transformable patterns to realize adversary mismatch and target person impersonation in physical world. The extensive evaluations demonstrate the vulnerability of deep re-ID systems to our attacks. 

\section*{Acknowledgments}
This work was supported in part by National Natural Science Foundation of China (Grants No. 61872274, 61822207 and U1636219), Equipment Pre-Research Joint Fund of Ministry of Education of China (Youth Talent) (Grant No. 6141A02033327), and Natural Science Foundation of Hubei Province (Grants No. 2017CFB503, 2017CFA047), and Fundamental Research Funds for the Central Universities (Grants No. 2042019gf0098, 2042018gf0043).

{\small
\bibliographystyle{ieee_fullname}
\bibliography{paperbib}

\begin{thebibliography}{10}\itemsep=-1pt

\bibitem{ahmed2015improved}
Ejaz Ahmed, Michael Jones, and Tim~K Marks.
\newblock An improved deep learning architecture for person re-identification.
\newblock In {\em Proc. of IEEE CVPR}, pages 3908--3916, 2015.

\bibitem{athalye2017synthesizing}
Anish Athalye and Ilya Sutskever.
\newblock Synthesizing robust adversarial examples.
\newblock {\em arXiv:1707.07397}, 2017.

\bibitem{carlini2017towards}
Nicholas Carlini and David Wagner.
\newblock Towards evaluating the robustness of neural networks.
\newblock In {\em Proc. of IEEE S\&P}, pages 39--57, 2017.

\bibitem{chen2016deep}
Shi-Zhe Chen, Chun-Chao Guo, and Jian-Huang Lai.
\newblock Deep ranking for person re-identification via joint representation
  learning.
\newblock {\em IEEE Transactions on Image Processing}, 25(5):2353--2367, 2016.

\bibitem{chen2017multi}
Weihua Chen, Xiaotang Chen, Jianguo Zhang, and Kaiqi Huang.
\newblock A multi-task deep network for person re-identification.
\newblock In {\em Proc. of AAAI}, pages 3988--3994, 2017.

\bibitem{cheng2016person}
De Cheng, Yihong Gong, Sanping Zhou, Jinjun Wang, and Nanning Zheng.
\newblock Person re-identification by multi-channel parts-based cnn with
  improved triplet loss function.
\newblock In {\em Proc. of IEEE CVPR}, pages 1335--1344, 2016.

\bibitem{ding2015deep}
Shengyong Ding, Liang Lin, Guangrun Wang, and Hongyang Chao.
\newblock Deep feature learning with relative distance comparison for person
  re-identification.
\newblock {\em Pattern Recognition}, 48(10):2993--3003, 2015.

\bibitem{eykholt2018robust}
Kevin Eykholt, Ivan Evtimov, Earlence Fernandes, Bo Li, Amir Rahmati, Chaowei
  Xiao, Atul Prakash, Tadayoshi Kohno, and Dawn Song.
\newblock Robust physical-world attacks on deep learning visual classification.
\newblock In {\em Proc. of IEEE CVPR}, pages 1625--1634, 2018.

\bibitem{gong2014re}
Shaogang Gong, Marco Cristani, Chen~Change Loy, and Timothy~M Hospedales.
\newblock The re-identification challenge.
\newblock In {\em Person re-identification}, pages 1--20. 2014.

\bibitem{goodfellow6572explaining}
Ian~J Goodfellow, Jonathon Shlens, and Christian Szegedy.
\newblock Explaining and harnessing adversarial examples.
\newblock {\em arXiv:1412.6572}.

\bibitem{grosse2016adversarial}
Kathrin Grosse, Nicolas Papernot, Praveen Manoharan, Michael Backes, and
  Patrick McDaniel.
\newblock Adversarial perturbations against deep neural networks for malware
  classification.
\newblock {\em arXiv:1606.04435}, 2016.

\bibitem{he2016deep}
Kaiming He, Xiangyu Zhang, Shaoqing Ren, and Jian Sun.
\newblock Deep residual learning for image recognition.
\newblock In {\em Proc. of IEEE CVPR}, pages 770--778, 2016.

\bibitem{kos2018adversarial}
Jernej Kos, Ian Fischer, and Dawn Song.
\newblock Adversarial examples for generative models.
\newblock In {\em Proc. of IEEE SPW}, pages 36--42, 2018.

\bibitem{krizhevsky2012imagenet}
Alex Krizhevsky, Ilya Sutskever, and Geoffrey~E Hinton.
\newblock Imagenet classification with deep convolutional neural networks.
\newblock In {\em Proc. of NIPS}, pages 1097--1105, 2012.

\bibitem{kurakin2016adversarial}
Alexey Kurakin, Ian Goodfellow, and Samy Bengio.
\newblock Adversarial examples in the physical world.
\newblock {\em arXiv:1607.02533}, 2016.

\bibitem{li2014feature}
Bo Li and Yevgeniy Vorobeychik.
\newblock Feature cross-substitution in adversarial classification.
\newblock In {\em Proc. of NIPS}, pages 2087--2095, 2014.

\bibitem{li2015scalable}
Bo Li and Yevgeniy Vorobeychik.
\newblock Scalable optimization of randomized operational decisions in
  adversarial classification settings.
\newblock In {\em Proc. of AISTATS}, pages 599--607, 2015.

\bibitem{li2014deepreid}
Wei Li, Rui Zhao, Tong Xiao, and Xiaogang Wang.
\newblock Deepreid: Deep filter pairing neural network for person
  re-identification.
\newblock In {\em Proc. of IEEE CVPR}, pages 152--159, 2014.

\bibitem{liu2016delving}
Yanpei Liu, Xinyun Chen, Chang Liu, and Dawn Song.
\newblock Delving into transferable adversarial examples and black-box attacks.
\newblock {\em arXiv:1611.02770}, 2016.

\bibitem{loy2009multi}
Chen~Change Loy, Tao Xiang, and Shaogang Gong.
\newblock Multi-camera activity correlation analysis.
\newblock In {\em Proc. of IEEE CVPR}, pages 1988--1995, 2009.

\bibitem{mahendran2015understanding}
Aravindh Mahendran and Andrea Vedaldi.
\newblock Understanding deep image representations by inverting them.
\newblock In {\em Proc. of IEEE CVPR}, pages 5188--5196, 2015.

\bibitem{moosavi2016deepfool}
Seyed-Mohsen Moosavi-Dezfooli, Alhussein Fawzi, and Pascal Frossard.
\newblock Deepfool: a simple and accurate method to fool deep neural networks.
\newblock In {\em Proc. of IEEE CVPR}, pages 2574--2582, 2016.

\bibitem{papernot2016transferability}
Nicolas Papernot, Patrick McDaniel, and Ian Goodfellow.
\newblock Transferability in machine learning: from phenomena to black-box
  attacks using adversarial samples.
\newblock {\em arXiv:1605.07277}, 2016.

\bibitem{papernot2016limitations}
Nicolas Papernot, Patrick McDaniel, Somesh Jha, Matt Fredrikson, Z~Berkay
  Celik, and Ananthram Swami.
\newblock The limitations of deep learning in adversarial settings.
\newblock In {\em Proc. of IEEE EuroS\&P}, pages 372--387, 2016.

\bibitem{sharif2016accessorize}
Mahmood Sharif, Sruti Bhagavatula, Lujo Bauer, and Michael~K Reiter.
\newblock Accessorize to a crime: Real and stealthy attacks on state-of-the-art
  face recognition.
\newblock In {\em Proc. of ACM CCS}, pages 1528--1540, 2016.

\bibitem{simonyan2014very}
Karen Simonyan and Andrew Zisserman.
\newblock Very deep convolutional networks for large-scale image recognition.
\newblock {\em arXiv:1409.1556}, 2014.

\bibitem{szegedy2015going}
Christian Szegedy, Wei Liu, Yangqing Jia, Pierre Sermanet, Scott Reed, Dragomir
  Anguelov, Dumitru Erhan, Vincent Vanhoucke, and Andrew Rabinovich.
\newblock Going deeper with convolutions.
\newblock In {\em Proc. of IEEE CVPR}, pages 1--9, 2015.

\bibitem{szegedy2013intriguing}
Christian Szegedy, Wojciech Zaremba, Ilya Sutskever, Joan Bruna, Dumitru Erhan,
  Ian Goodfellow, and Rob Fergus.
\newblock Intriguing properties of neural networks.
\newblock {\em arXiv:1312.6199}, 2013.

\bibitem{wang2016joint}
Faqiang Wang, Wangmeng Zuo, Liang Lin, David Zhang, and Lei Zhang.
\newblock Joint learning of single-image and cross-image representations for
  person re-identification.
\newblock In {\em Proc. of IEEE CVPR}, pages 1288--1296, 2016.

\bibitem{wang2013intelligent}
Xiaogang Wang.
\newblock Intelligent multi-camera video surveillance: A review.
\newblock {\em Pattern recognition letters}, 34(1):3--19, 2013.

\bibitem{xiao2016learning}
Tong Xiao, Hongsheng Li, Wanli Ouyang, and Xiaogang Wang.
\newblock Learning deep feature representations with domain guided dropout for
  person re-identification.
\newblock In {\em Proc. of IEEE CVPR}, pages 1249--1258, 2016.

\bibitem{yi2014deep}
Dong Yi, Zhen Lei, Shengcai Liao, and Stan~Z Li.
\newblock Deep metric learning for person re-identification.
\newblock In {\em Proc. of IEEE ICPR}, pages 34--39, 2014.

\bibitem{yu2013harry}
Shoou-I Yu, Yi Yang, and Alexander Hauptmann.
\newblock Harry potter's marauder's map: Localizing and tracking multiple
  persons-of-interest by nonnegative discretization.
\newblock In {\em Proc. of IEEE CVPR}, pages 3714--3720, 2013.

\bibitem{zhang2017age}
Zhifei Zhang, Yang Song, and Hairong Qi.
\newblock Age progression/regression by conditional adversarial autoencoder.
\newblock In {\em Proc. of IEEE CVPR}, pages 5810--5818, 2017.

\bibitem{zhang2019image}
Zhifei Zhang, Zhaowen Wang, Zhe Lin, and Hairong Qi.
\newblock Image super-resolution by neural texture transfer.
\newblock In {\em Proc. of IEEE CVPR}, pages 7982--7991, 2019.

\bibitem{zheng2016person}
Liang Zheng, Yi Yang, and Alexander~G Hauptmann.
\newblock Person re-identification: Past, present and future.
\newblock {\em arXiv:1610.02984}, 2016.

\bibitem{zheng2017discriminatively}
Zhedong Zheng, Liang Zheng, and Yi Yang.
\newblock A discriminatively learned cnn embedding for person reidentification.
\newblock {\em ACM Transactions on Multimedia Computing, Communications, and
  Applications (TOMM)}, 14(1):13, 2017.

\bibitem{zhong2018camera}
Zhun Zhong, Liang Zheng, Zhedong Zheng, Shaozi Li, and Yi Yang.
\newblock Camera style adaptation for person re-identification.
\newblock In {\em Proc. of IEEE CVPR}, pages 5157--5166, 2018.

\end{thebibliography}
}

\end{document}